\documentclass[lettersize,journal]{IEEEtran}  

\IEEEoverridecommandlockouts                              

\usepackage{amsmath,amssymb,amsfonts}
\usepackage{algorithm}
\usepackage{algorithmic}
\usepackage{graphicx}
\usepackage{textcomp}
\usepackage{xcolor}
\usepackage{multirow}
\usepackage{booktabs}
\usepackage{bm}
\usepackage{mathrsfs}
\usepackage{float}
\usepackage{subfigure}
\usepackage{marvosym}
\usepackage{hyperref}

\title{\LARGE \bf
Underwater Embodied Intelligence for Autonomous Robots: A Constraint-Coupled Perspective on Planning, Control, and Deployment}

\author{Jingzehua Xu$^{1}$, \IEEEmembership{Student Member, IEEE}, Guanwen Xie$^{2}$, \IEEEmembership{Student Member, IEEE},\\Jiwei Tang$^{3}$, \IEEEmembership{Student Member, IEEE}, Shuai Zhang$^{4}$, \IEEEmembership{Member, IEEE}, and Xiaofan Li$^{1,}\textsuperscript{\Letter}$
\thanks{$^{1}$J. Xu and X. Li are with Department of Mechanical Engineering, The University of Hong Kong, Pokfulam Road, Hong Kong, China; E-mail: xjzh23@mails.tsinghua.edu.cn, lixf@hku.hk.}
\thanks{$^{2}$G. Xie is with Tsinghua Shenzhen International Graduate School, Tsinghua University, Shenzhen, 518055, China. E-mail: xgw24@mails.tsinghua.edu.cn.}%
\thanks{$^{3}$J. Tang is with Department of Data and Systems Engineering, The University of Hong Kong, Pokfulam, Hong Kong, E-mail: tangjiwei@connect.hku.hk.}
\thanks{$^{4}$S. Zhang is with Department of Data Science, New Jersey Institute of Technology, NJ 07102, USA. E-mail: sz457@njit.edu.}
\thanks{$\textsuperscript{\Letter}$ Corresponding author.}
}
\begin{document}

\markboth{}%
{Shell \MakeLowercase{\textit{et al.}}: A Sample Article Using IEEEtran.cls for IEEE Journals}

\maketitle

\begin{abstract}
Autonomous underwater robots are increasingly deployed for environmental monitoring, infrastructure inspection, subsea resource exploration, and long-horizon exploration. Yet, despite rapid advances in learning-based planning and control, reliable autonomy in real ocean environments remains fundamentally constrained by tightly coupled physical limits. Hydrodynamic uncertainty, partial observability, bandwidth-limited communication, and energy scarcity are not independent challenges; they interact within the closed perception–planning–control loop and often amplify one another over time. This Review develops a constraint-coupled perspective on underwater embodied intelligence, arguing that planning and control must be understood within tightly coupled sensing, communication, coordination, and resource constraints in real ocean environments. We synthesize recent progress in reinforcement learning, belief-aware planning, hybrid control, multi-robot coordination, and foundation-model integration through this embodied perspective. Across representative application domains, we show how environmental monitoring, inspection, exploration, and cooperative missions expose distinct stress profiles of cross-layer coupling. To unify these observations, we introduce a cross-layer failure taxonomy spanning epistemic, dynamic, and coordination breakdowns, and analyze how errors cascade across autonomy layers under uncertainty. Building on this structure, we outline research directions toward physics-grounded world models, certifiable learning-enabled control, communication-aware coordination, and deployment-aware system design. By internalizing constraint coupling rather than treating it as an external disturbance, underwater embodied intelligence may evolve from performance-driven adaptation toward resilient, scalable, and verifiable autonomy under real ocean conditions.
\end{abstract}

\section{Introduction}

\begin{figure*}[!t]
    \centering
    \includegraphics[width=1.0\linewidth]{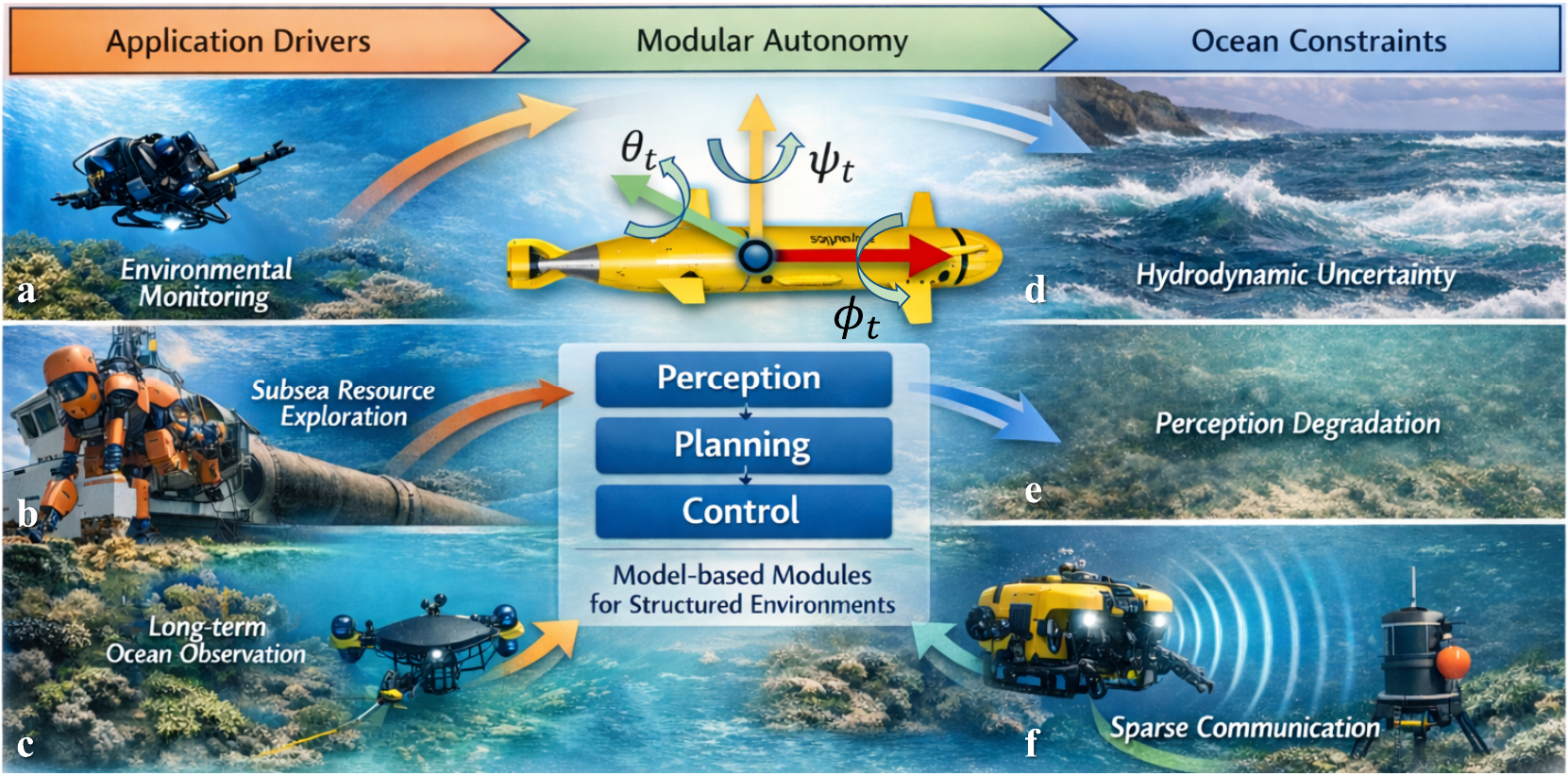}
    \caption{\textbf{Application drivers and environmental constraints shaping underwater robotic autonomy}. Representative application scenarios motivating the deployment of autonomous underwater robots include environmental monitoring, subsea resource exploration, and long-term ocean observation. These tasks require persistent operation, large-area coverage, and reduced human supervision. (a) Environmental monitoring of marine ecosystems using autonomous underwater vehicles \cite{1}. (b) Subsea resource exploration and infrastructure inspection in complex underwater environments \cite{7}. (c) Long-term ocean observation requiring sustained and wide-area data collection \cite{147}. To support such missions, most existing systems adopt a modular autonomy architecture integrating perception, planning, and control for environment understanding and decision-making. However, reliable autonomy in real oceans remains challenged by several environmental constraints: (d) hydrodynamic uncertainty from time-varying currents and fluid–vehicle interactions; (e) perception degradation due to turbidity, light attenuation, and sensing limitations; and (f) sparse communication caused by bandwidth-limited and latency-prone acoustic channels.}
    \label{fig:1}
\end{figure*}

\IEEEPARstart{U}{nderwater} robots are increasingly central to maritime applications including environmental monitoring \cite{1,2,3}, offshore infrastructure inspection \cite{4,5,6}, subsea resource exploration \cite{7,8}, and long-term ocean observation \cite{9,147}. Compared with human-operated or remotely controlled systems, underwater robots enable persistent deployment, reduce operational costs, and access hazardous or otherwise unreachable environments \cite{11,12,13}. As mission requirements expand toward longer durations, larger spatial coverage, and reduced human supervision, autonomy shifts from an optional capability to a foundational requirement \cite{14,15,16}. Consequently, the next generation of underwater robots must operate reliably under limited sensing, uncertain dynamics, intermittent communication, and constrained onboard resources \cite{17,18,19} (Fig.~\ref{fig:1}a-c).

However, achieving such reliability in real ocean environments remains fundamentally challenging. Traditional underwater autonomy has largely relied on modular pipelines combining model-based planning, state estimation, and feedback control \cite{20,21,22}. When system dynamics are well characterized and sensing remains reliable, these approaches provide strong performance and formal guarantees \cite{23,24,25}. In practice, though, realistic ocean environments rarely satisfy these assumptions \cite{26}. Time-varying currents introduce disturbances that are difficult to model precisely; optical perception degrades under turbidity and light attenuation; acoustic sensing and communication are constrained by low bandwidth and latency \cite{27,28,29,30}. Furthermore, hydrodynamic models depend strongly on vehicle geometry, operating regime, and environmental conditions, limiting generalizability across platforms and deployment sites \cite{31,32}. Taken together, these factors expose a structural gap between carefully engineered autonomy stacks and real-world deployment, where robustness, adaptivity, and long-horizon reliability become dominant bottlenecks \cite{33} (Fig.~\ref{fig:1}d-f).

Against this backdrop, a paradigm of underwater embodied intelligence has emerged to rethink autonomy under ocean constraints \cite{34,35,36}. Rather than treating perception, planning, and control as loosely coupled algorithmic modules, embodied intelligence views autonomy as arising from closed-loop interaction among a robot’s physical embodiment, its sensing and actuation mechanisms, and the surrounding marine environment \cite{37,38}. Within this framework, learning-based approaches, including reinforcement learning and imitation learning, are increasingly used to acquire adaptive policies through interaction \cite{39,40}. More recently, language-based and vision–language models have extended autonomy toward higher-level semantic reasoning and task abstraction \cite{41,42}. Importantly, in underwater domains, these capabilities cannot operate independently of physical feasibility: hydrodynamics, sensing quality, communication bandwidth, and energy availability are tightly coupled constraints that reshape the perception–planning–control loop \cite{43,44,45,46}. Intelligence in this context is therefore not merely computational sophistication, but physically grounded, uncertainty-aware interaction within a dynamically coupled medium.

Building upon this shift in perspective, this review advances a unifying thesis: underwater autonomy should be understood as a constraint-coupled embodied systems architecture, in which learning, sensing, control, and communication are co-designed under tightly coupled physical limits rather than appended as loosely connected algorithmic layers. Importantly, this perspective does not reject hierarchical structure in autonomy architectures; instead, it highlights the limitations of loosely coupled modular pipelines in which perception, planning, and control are optimized largely in isolation, and emphasizes tightly integrated hierarchies grounded in shared representations of uncertainty, dynamics, and resource constraints. From this standpoint, mission progress, uncertainty regulation, and dynamic feasibility are inseparable objectives. Planning decisions influence observability and controllability; control actions affect energy consumption and sensing stability; communication constraints reshape coordination and belief consistency. Robust underwater autonomy, therefore, requires regulating cross-layer error propagation across the perception–planning–control loop under communication and resource constraints while preserving physical feasibility and operational endurance.

This conceptual reframing is also reflected in the historical evolution of the field. Over the past two decades, underwater autonomy has undergone a gradual structural transition. Early systems emphasized model-based planning and control under structured assumptions \cite{58, 102}. Subsequent developments incorporated learning-based augmentation to compensate for model mismatch and environmental disturbances \cite{109}. More recently, reinforcement learning, belief-aware planning, and foundation-model-based reasoning have shifted attention toward closed-loop, uncertainty-aware system integration \cite{55, 56}. In essence, this evolution marks a deeper shift from modular autonomy pipelines toward embodied architectures that explicitly co-design perception, adaptation, and control under environmental and resource constraints \cite{42}.

Guided by this perspective, this review first clarifies the defining characteristics of underwater embodied intelligence and explains why embodiment in marine systems is structurally distinct from terrestrial or aerial robotics. We then introduce a systems-level abstraction that interprets underwater autonomy as a constraint-coupled regulation process over joint state, belief, and resource spaces and examine its implications for planning and control. Finally, we analyze representative application domains, identify cross-layer failure mechanisms, and outline future research directions toward robust and scalable ocean autonomy.

\section{An Overview of Underwater Embodied Intelligence}

\subsection{Definition and Scope}

Building upon the constraint-coupled perspective introduced in Section I, we begin by clarifying what is meant by underwater embodied intelligence. In this review, underwater embodied intelligence refers to autonomy paradigms in which perception, planning, and control policies are jointly learned, adapted, and deployed in tight coupling with the robot’s physical embodiment and the surrounding ocean environment, rather than being designed as loosely connected algorithmic modules \cite{47,48,49}. 

This clarification is necessary because, in underwater domains, embodiment is not a secondary implementation detail but a dominant structural constraint. Hydrodynamic forces couple translational and rotational dynamics; buoyancy and restoring moments depend on geometry and mass distribution; and sensing quality varies sharply with turbidity, illumination, and acoustic multipath \cite{50,51}. As a consequence, algorithmic performance cannot be evaluated independently of what the vehicle can reliably perceive, execute, and sustain in situ \cite{52,53}. In other words, autonomy is fundamentally conditioned by physical feasibility and environmental uncertainty, rather than determined solely by computational capability.

To make this perspective operational rather than purely conceptual, we specify concrete criteria that characterize underwater embodied intelligence. In this review, we consider approaches as relevant to underwater embodied intelligence if they satisfy at least one of the following conditions \cite{54}:  
(i) explicitly modeling ocean-induced uncertainty or partial observability within planning or control;  
(ii) treating sensing as an active, uncertainty-reducing process rather than passive perception;  
(iii) embedding embodiment constraints such as actuation limits, energy budgets, or platform-specific hydrodynamics directly into decision-making; or  
(iv) providing a practical pathway to deployment under distribution shift, for example through sim-to-real transfer, system identification, online adaptation, or safety-aware supervision.
These criteria provide a pragmatic inclusion boundary for the literature surveyed in this review. However, the strongest form of underwater embodied intelligence arises when these aspects are integrated within a closed-loop autonomy architecture that jointly regulates state evolution, belief consistency, and embodiment constraints. It is precisely this tighter systems-level coupling that motivates the constraint-coupled perspective developed in the following sections.

By contrast, approaches that neglect these structural constraints are less central to the embodied perspective emphasized in this review. In particular, purely data-driven policies trained in simulation should be regarded as only weakly embodied unless uncertainty, embodiment constraints, or deployment conditions are explicitly incorporated. Similarly, modular pipelines that simply append learning components without integrating physical feasibility into decision-making only partially realize the embodied paradigm. 

In aggregate, the defining characteristic of underwater embodied intelligence lies in the explicit co-design of learning, uncertainty representation, and embodiment constraints within a closed-loop autonomy framework. It is this integration, rather than the mere use of learning alone, that distinguishes embodied systems from conventional algorithmic pipelines.

\begin{figure*}[!t]
    \centering
    \includegraphics[width=1.0\linewidth]{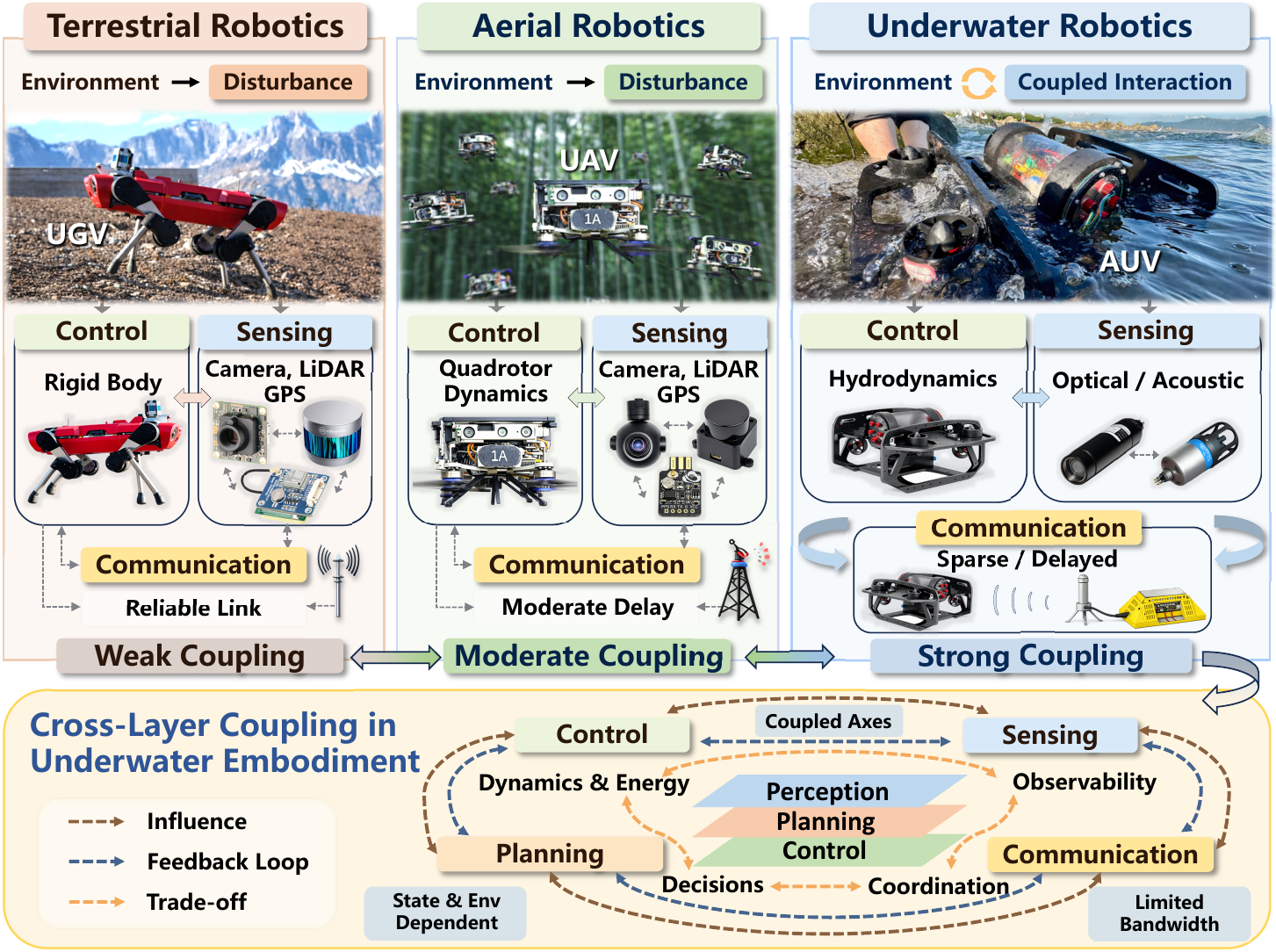}
    \caption{\textbf{Structural distinctiveness of underwater embodiment compared to terrestrial and aerial systems.} In terrestrial and aerial robotics, environmental interaction primarily acts as external disturbance \cite{148,149}. By contrast, underwater systems operate within a dense, viscous, and dynamically coupled fluid medium that simultaneously reshapes controllability, observability, and communication reliability \cite{152}. Hydrodynamic effects couple translational and rotational dynamics; optical and acoustic sensing quality depends on motion and environmental conditions; and acoustic communication introduces sparse, delayed coordination. These factors create recursive cross-layer coupling across perception, planning, and control while also constraining coordination through sparse and delayed communication. As a result, autonomy in underwater systems cannot be decomposed into independent modules without structural fragility.}
    \label{fig:2}
\end{figure*}

\subsection{Why Underwater Embodiment is Structurally Distinct}

Having clarified the scope and defining characteristics of underwater embodied intelligence, we next examine why embodiment in underwater systems is structurally distinct from its counterparts in terrestrial or aerial robotics. While embodied intelligence has been widely studied across robotic domains, the ocean environment introduces a qualitatively different form of physical coupling that reshapes the autonomy problem at a systems level \cite{150, 151} (Fig.~\ref{fig:2}).

First of all, the surrounding medium fundamentally alters vehicle dynamics. Unlike ground vehicles operating on relatively rigid surfaces or aerial robots navigating in comparatively weakly coupled air flows, underwater vehicles continuously interact with a dense, viscous, and often turbulent fluid medium \cite{148,149,152}. This interaction is not merely a source of disturbance; rather, it directly modifies controllability, stability margins, and energy expenditure \cite{80}. Hydrodynamic effects introduce added mass, nonlinear drag, and cross-axis coupling, such that motion along one axis can induce responses along others \cite{32}. As a consequence, dynamic feasibility becomes inseparable from environmental conditions, and control authority depends explicitly on the state of the surrounding fluid.

Beyond dynamics, the same medium simultaneously reshapes perception. Optical sensing degrades rapidly with turbidity and light attenuation, while acoustic sensing is affected by multipath interference, limited bandwidth, and latency \cite{96}. Therefore, observability is neither static nor uniform; it is state-dependent and environment-dependent. What the robot can perceive reliably is tightly coupled to how it moves and where it positions itself \cite{64}. In effect, motion decisions influence sensing quality, and sensing limitations constrain motion planning \cite{151}. Perception and motion therefore interact within the broader perception–planning–control loop rather than forming independent algorithmic layers.

In addition to dynamics and sensing, communication introduces a further dimension of coupling. Underwater acoustic communication provides limited data rates and significant delays, which fragment shared belief states in multi-robot systems \cite{17}. Consequently, coordination strategies must explicitly account for asynchronous updates, partial information, and intermittent connectivity \cite{171}. Unlike many terrestrial multi-agent systems that assume stable and high-bandwidth communication, underwater cooperation is fundamentally shaped by communication physics and resource constraints \cite{146}.

Taken as a whole, these factors reveal a defining structural characteristic: the surrounding medium simultaneously influences dynamics, sensing, and coordination, and these influences propagate across the perception–planning–control loop. Disturbances in one layer can cascade into others. For example, sensing degradation may bias planning decisions, which in turn push controllers toward energetically expensive or dynamically unstable regimes, further degrading estimation and coordination \cite{78}. Such cross-layer interactions accumulate over long horizons, amplifying uncertainty and resource trade-offs.

For this reason, underwater autonomy cannot be decomposed into fully independent perception, planning, and control modules without incurring structural fragility. Underwater embodied intelligence should therefore not be interpreted as a simple domain transfer of generic embodied AI methods. Rather, it represents a systems-level response to tightly coupled physical constraints in which controllability, observability, communication reliability, and energy sustainability co-evolve.

This structural distinctiveness, in turn, motivates the need for a unified systems abstraction that explicitly captures these couplings. We introduce such an abstraction in the following subsection.

\begin{figure*}[!t]
    \centering
    \includegraphics[width=1.0\linewidth]{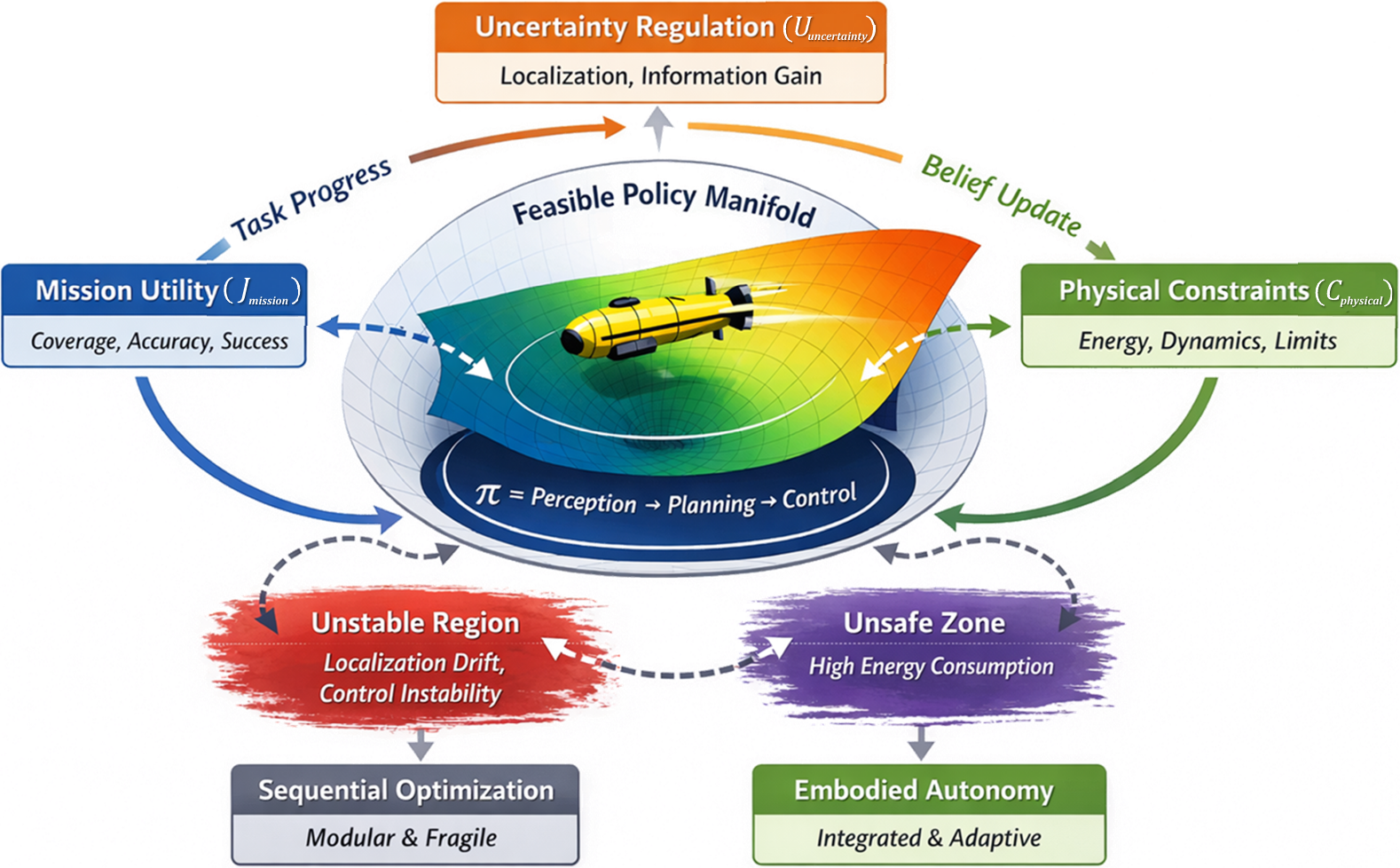}
    \caption{\textbf{Embodied autonomy as constraint-coupled optimization in underwater systems.} Rather than optimizing perception, planning, and control as isolated modules, underwater autonomy is more appropriately understood as a closed-loop, multi-objective regulation process over joint state, belief, and resource spaces. Mission utility, uncertainty regulation, and physical feasibility are structurally coupled: actions that improve task progress may increase localization drift, energy consumption, or dynamic instability, while uncertainty reduction and safety preservation can in turn reshape effective mission performance. The feasible policy manifold therefore represents the set of embodied strategies that balance these competing yet interdependent objectives, whereas sequential modular optimization is more likely to drive the system toward unstable or unsafe regimes under realistic ocean conditions.}
    \label{fig:3}
\end{figure*}

\subsection{A Systems Abstraction: Embodied Autonomy as Constraint-Coupled Optimization}

The structural couplings discussed in the previous subsection indicate that underwater autonomy cannot be fully explained by analyzing perception, planning, and control in isolation. Rather, a unifying systems abstraction is required to describe how mission objectives, uncertainty regulation, and embodiment constraints interact within a single closed loop.

From this standpoint, underwater embodied intelligence can be interpreted as a constrained multi-objective optimization process evolving over joint state, belief, and resource spaces. Let $\pi$ denote a closed-loop policy that integrates perception, planning, and control within the autonomy stack. To illustrate this idea conceptually, a simplified schematic abstraction of embodied autonomy may therefore be written as
\begin{equation}
\min_{\pi}
\mathbb{E}[J_{\text{mission}}]
+ \lambda_1 \mathbb{E}[U_{\text{uncertainty}}]
+ \lambda_2 \mathbb{E}[C_{\text{physical}}],
\end{equation}
where $J_{\text{mission}}$ represents task-level utility such as coverage efficiency, inspection accuracy, or tracking success; $U_{\text{uncertainty}}$ captures epistemic regulation including localization stability, belief consistency, or information gain; and $C_{\text{physical}}$ encodes embodiment-related costs such as actuation limits, dynamic feasibility margins, and energy consumption. Communication constraints primarily influence the uncertainty term through delayed information propagation and fragmented belief states across agents. They also interact with embodiment constraints through bandwidth limitations and energy expenditure. For clarity, these effects are incorporated implicitly into the uncertainty and physical cost terms, while communication remains a distinct structural factor shaping coordination and scalability in multi-robot systems.

At first glance, this formulation may appear similar to conventional multi-objective optimization \cite{197}. However, the crucial distinction lies in the fact that these objectives are inherently interdependent rather than separable. Actions that maximize immediate mission utility may increase localization drift or destabilize belief estimates \cite{161}. Conversely, aggressive information-seeking maneuvers may reduce epistemic uncertainty while significantly increasing energy expenditure or pushing the system toward dynamically unstable regimes \cite{182}. Even conservative stabilization strategies can degrade sensing resolution or prolong mission duration \cite{156}. 

As a result, embodied autonomy emerges not as the optimization of independent sub-goals, but as a continuous negotiation among task advancement, uncertainty regulation, and physical feasibility. The policy must internalize the structural couplings identified earlier because each action simultaneously affects mission progress, belief evolution, and embodiment constraints (Fig.~\ref{fig:3}).

This perspective also clarifies why classical modular autonomy pipelines often degrade under realistic ocean conditions. In many traditional architectures, mission objectives are optimized first, while uncertainty handling and physical constraints are incorporated as secondary corrections or safety layers \cite{134}. In underwater systems, however, uncertainty and embodiment constraints reshape the effective objective landscape itself \cite{151}. Planning decisions influence future observability; control actions affect sensing stability and energy reserves; communication limitations alter belief consistency across agents \cite{79,125,93}. Consequently, sequential optimization fails to capture cross-layer dependencies that accumulate over long horizons.

Moreover, the optimization problem is inherently dynamic and context-dependent. The relative weighting coefficients $\lambda_1$ and $\lambda_2$ are not fixed parameters but implicitly vary with environmental conditions, mission phase, and resource availability. During early exploration in unknown terrain, uncertainty reduction may dominate decision-making \cite{159}. During extended monitoring missions, energy preservation may become a primary concern \cite{154}. Near complex infrastructure, dynamic safety margins may override coverage efficiency \cite{166}. Embodied autonomy, therefore, operates as a context-sensitive regulation mechanism rather than a static objective minimization.

Viewed through this lens, intelligence in underwater systems cannot be reduced to maximizing task rewards alone. Instead, it reflects the capacity to maintain stable belief states, preserve controllability, and allocate limited resources while progressing toward mission goals. The system must prevent cross-layer error propagation across the perception–planning–control loop so that disturbances in one component do not cascade into systemic failure.

With this unified abstraction established, we next examine how these constraint-coupled principles manifest concretely in the domains of planning and control.

\subsection{Underwater Embodied Intelligence for Planning of Autonomous Robots}

With the constraint-coupled systems abstraction established above, we now examine how these principles manifest concretely in underwater planning. Under the embodied perspective, planning is no longer confined to geometric path generation. Instead, it becomes a regulation process over joint state, belief, and resource spaces, where each decision simultaneously affects mission progress, uncertainty evolution, and physical feasibility.

To begin with, underwater planning is fundamentally shaped by partial observability. Acoustic positioning systems and dead reckoning provide intermittent updates and accumulate drift over long horizons \cite{62,63}. Consequently, the planner must reason not only about physical state evolution but also about belief dynamics. Rather than assuming accurate localization, embodied planning increasingly adopts belief-aware or history-conditioned formulations in which task advancement and uncertainty regulation are optimized jointly \cite{64,65,66}. In practice, this includes recurrent policies, memory-augmented architectures, and receding-horizon strategies that balance exploitation with information acquisition \cite{67,68,69,70}. As a result, planning performance is evaluated not only through geometric metrics such as path length or time-to-goal, but also through uncertainty-aware indicators including localization error growth, robustness under delayed sensing, and success probability under observation degradation \cite{71,72,73}.

Notably, uncertainty regulation in underwater planning is not merely a passive correction mechanism. Instead, perception becomes an actively regulated resource. In many underwater missions, such as infrastructure inspection or ecological monitoring, the overall mission value depends directly on the quality of perception \cite{74,75,38}. High-resolution sensing may require specific viewpoints, stand-off distances, or motion profiles that mitigate turbidity effects or sonar artifacts \cite{75,76,77}. Therefore, planning must explicitly couple trajectory selection with observability optimization \cite{78,79,80}. Typical behaviors include revisiting uncertain regions to stabilize belief states, adapting altitude or velocity to maintain measurement fidelity, and approaching structures from geometrically informative angles \cite{81,82}. At the same time, these information-seeking maneuvers remain constrained by actuation limits and hydrodynamic feasibility \cite{83,84}. The planner must therefore continuously negotiate between information gain and dynamic stability.

Beyond geometry and sensing, task-level abstraction introduces an additional layer of coupling. Many underwater missions require semantic prioritization, such as identifying defect signatures, habitat boundaries, or specific targets \cite{85,34}. Language-based and vision–language models increasingly support higher-level reasoning by mapping sensory observations to semantic hypotheses and generating structured subgoals \cite{87,35}. However, semantic intent must ultimately be grounded in dynamically feasible motion. Without embodiment-aware constraints, high-level goals risk producing infeasible or unsafe trajectories. Consequently, hierarchical planning architectures are often adopted, where semantic reasoning operates at an abstract layer while lower-level planners enforce feasibility, safety margins, and uncertainty regulation \cite{88,89}. This separation preserves interpretability while maintaining closed-loop grounding.

When planning extends to multi-robot systems, communication physics introduces yet another structural constraint. Underwater acoustic communication provides limited bandwidth and significant delay, precluding centralized, high-frequency coordination \cite{91,92}. As a result, embodied planning emphasizes decentralized strategies, sparse communication protocols, and policies that determine when and what information to transmit \cite{93}. Leader–follower structures, relay behaviors, and heterogeneity-aware task allocation reduce reliance on continuous connectivity \cite{94,95}. Meanwhile, communication decisions compete with propulsion and sensing for limited onboard energy \cite{97,98,99}. Planning therefore operates under a three-way coupling among mission objectives, belief synchronization, and energy sustainability within the broader perception–planning–control loop.

Considered together, these developments reveal that underwater embodied planning is not merely path optimization under constraints. Rather, it is a dynamic coordination process that jointly regulates mission progress, epistemic stability, physical feasibility, communication reliability, and resource allocation. Actions reshape future observability and controllability; sensing quality influences trajectory feasibility; communication constraints modify belief consistency across agents. Effective planning must therefore internalize cross-layer couplings to prevent uncertainty accumulation and resource imbalance from cascading into mission failure.

Having examined how embodied intelligence reshapes planning, we now turn to its implications for control, where similar constraint-coupled principles govern stabilization and disturbance rejection.

\subsection{Underwater Embodied Intelligence for Control of Autonomous Robots}

Having examined how embodied principles reshape planning, we now turn to control, where the same constraint-coupled dynamics govern stabilization and disturbance rejection. Whereas planning regulates future state–belief evolution, control determines whether those trajectories remain dynamically feasible when executed within the perception–planning–control loop. In this sense, control serves as the immediate realization layer of embodied autonomy.

Traditionally, underwater control has relied on model-based techniques such as PID variants, backstepping, sliding-mode control, and model predictive control \cite{103,104,108}. Under well-characterized dynamics, these approaches can provide stability guarantees and predictable performance. However, in practical deployments, modeling assumptions are frequently violated. Time-varying currents, unmodeled hydrodynamic couplings, actuator nonlinearities, and sensor noise often degrade controller performance \cite{39,106}. Moreover, in underwater systems, disturbance rejection and energy dissipation are tightly intertwined: aggressive corrective actions may preserve short-term stability but accelerate energy depletion or induce oscillatory dynamics over longer horizons. Thus, purely model-based control struggles to balance robustness and resource sustainability.

Against this backdrop, embodied intelligence integrates learning and adaptation directly into the control loop while preserving a physics-informed structure \cite{47,107}. Importantly, learning is typically introduced not as a replacement for classical control, but as a structured augmentation. Hybrid architectures, therefore, combine a stabilizing baseline controller that ensures nominal safety with learning-based components that compensate for model mismatch, disturbance variation, and environmental uncertainty \cite{109,110}. In many cases, structured policy learning constrains learned outputs to interpretable intermediate variables, such as desired angular rates, thrust allocation factors, or gain adjustments, which are subsequently mapped to actuator commands through established control laws \cite{111,112}. This layered design preserves physical consistency while enabling adaptive performance improvement.

Within such hybrid frameworks, reinforcement learning has become a prominent tool for enhancing robustness in attitude stabilization, depth regulation, and trajectory tracking under current-induced perturbations \cite{36,113}. To mitigate distribution shift between simulation and deployment, training often incorporates domain randomization over hydrodynamic coefficients, thruster dynamics, and sensor noise \cite{106,114}. Nevertheless, robustness cannot be assessed solely through nominal tracking accuracy. Instead, embodied control requires distribution-aware evaluation: how error tails behave under rare disturbances, how constraint violations evolve over extended missions, and whether actuator saturation or instability emerges under previously unseen environmental conditions \cite{115,116}. Accordingly, performance assessment must account for both stability margins and long-horizon reliability.

In addition to disturbance robustness, control must also explicitly address partial observability. As discussed in the planning context, underwater sensing can degrade abruptly due to turbidity, acoustic occlusions, or intermittent localization \cite{117,182}. Consequently, embodied control increasingly incorporates recurrent or belief-driven architectures that infer latent disturbances from temporal context \cite{27,119}. Such mechanisms enable proactive compensation rather than purely reactive correction. At the same time, learned control actions remain bounded by embodiment constraints, including inertia, added mass effects, actuator saturation limits, and finite energy budgets \cite{121,122}. Failure to respect these constraints may induce oscillations, destabilization, or rapid resource exhaustion.

Beyond robustness and observability, safety and verifiability constitute central open challenges. Learning-enabled controllers can exhibit unexpected behavior under distribution shift, particularly in long-duration missions or multi-robot coordination scenarios \cite{123}. To mitigate such risks, constraint-aware learning, action filtering, supervisory switching between learned and conservative controllers, and runtime monitoring mechanisms are increasingly incorporated \cite{124,125}. In this hierarchical structure, language-based or foundation-model components may influence high-level objectives or adaptive parameter tuning, yet low-level stabilization remains governed by physically grounded control laws. Clear role separation ensures that semantic abstraction enhances adaptability without destabilizing the dynamic safety envelope.

Altogether, underwater embodied control is not merely disturbance rejection augmented by learning. Rather, it is a closed-loop regulation process that balances robustness, uncertainty adaptation, physical feasibility, and resource sustainability under tightly coupled ocean constraints. Control actions influence future observability within the perception–planning–control loop, affect energy reserves, and shape coordination reliability across agents. Therefore, similar to planning, effective control must internalize cross-layer couplings to prevent local corrective decisions from cascading into systemic instability.

With both planning and control analyzed through the lens of constraint-coupled embodiment, we now turn to representative application domains, where these principles are stress-tested under real deployment conditions.

\section{Applications}
Having established the conceptual and systems foundations of underwater embodied intelligence, we next examine how these principles are stress-tested in real deployment contexts. Representative application domains act as structural probes that reveal how uncertainty, dynamic constraints, communication limitations, and resource restrictions interact within closed-loop autonomy systems centered on the perception–planning–control loop. Depending on mission type, the dominant tension may arise from uncertainty accumulation, proximity-driven safety constraints, long-horizon drift, or communication sparsity in multi-robot systems.

\subsection{Persistent Environmental Monitoring and Adaptive Sampling}

\begin{figure*}[!t]
    \centering
    \includegraphics[width=1.0\linewidth]{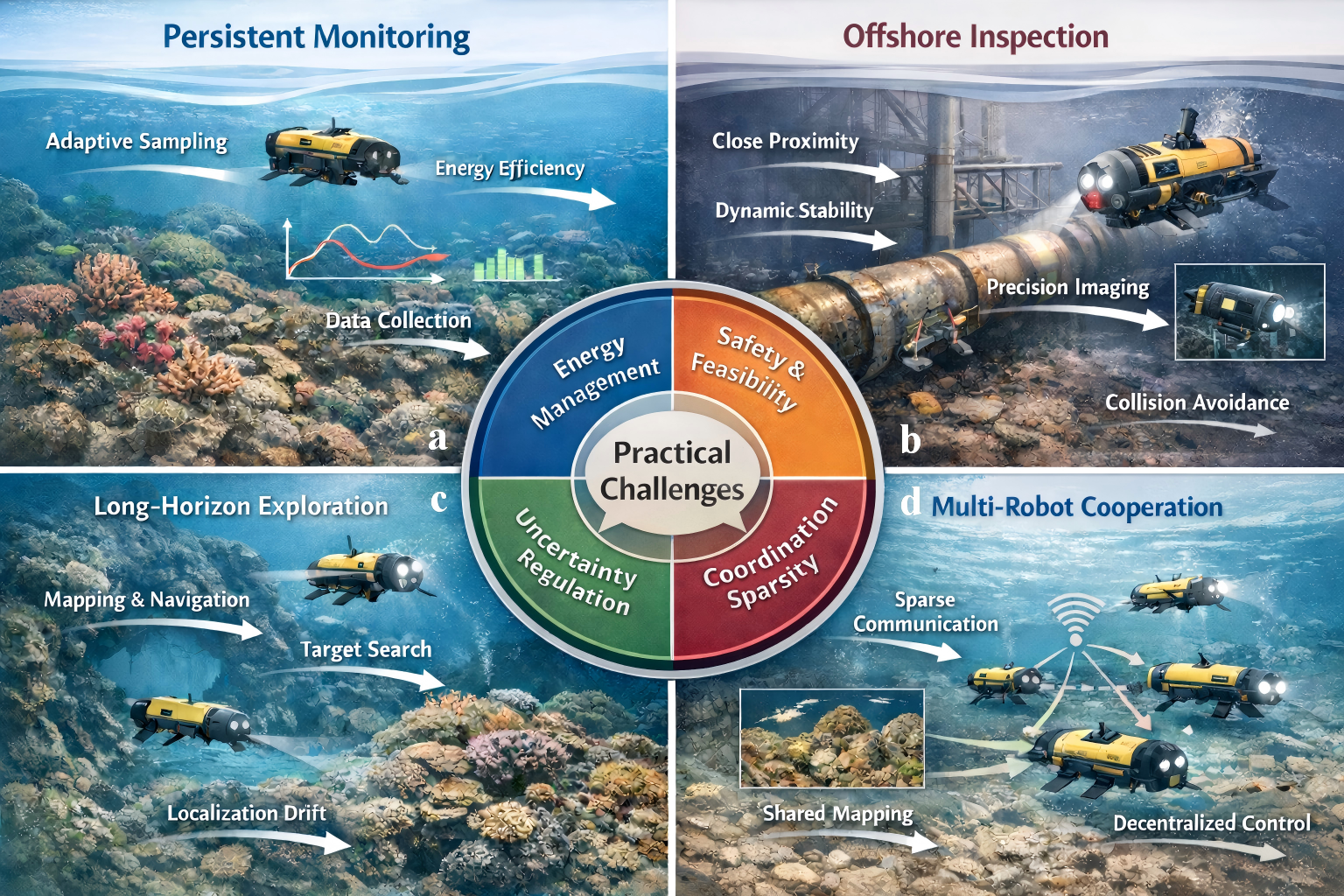}
    \caption{\textbf{Application domains as structural stress tests of embodied autonomy.} Representative underwater mission scenarios impose distinct yet partially overlapping constraint profiles that collectively challenge the stability and robustness of embodied autonomous systems. (a) Persistent monitoring places sustained pressure on long-horizon epistemic stability, requiring reliable state estimation and uncertainty management over extended deployments while simultaneously maintaining strict energy endurance constraints. (b) Infrastructure inspection emphasizes proximity-driven safety and dynamic feasibility, where operations near complex structures demand precise motion control, robust collision avoidance, and stable perception under turbulent hydrodynamic conditions. (c) Exploration magnifies the accumulation of epistemic uncertainty and the difficulty of maintaining globally consistent maps in previously unobserved environments, often under limited sensing and intermittent localization cues. (d) Multi-robot cooperation introduces additional system-level complexity through coordination sparsity, communication delays, and cross-agent belief fragmentation, requiring decentralized reasoning and bandwidth-aware information exchange. Although the dominant operational pressures differ across these domains, each scenario ultimately exposes the same underlying structural coupling across mission objectives, uncertainty regulation, physical feasibility, communication reliability, and long-term resource sustainability.}
    \label{fig:4}
\end{figure*}

Among underwater mission types, persistent environmental monitoring represents one of the most demanding long-duration autonomy scenarios \cite{2}. In these deployments, robots must operate over extended time horizons while coping with evolving environmental conditions, localization drift, and limited energy budgets \cite{84,140}. Consequently, monitoring tasks expose a defining structural coupling in underwater autonomy: epistemic uncertainty and mission endurance evolve together, and neither can be treated as secondary \cite{66} (Fig.~\ref{fig:4}a).

To understand this coupling more concretely, consider the shift from fixed waypoint execution to adaptive sampling \cite{61, 153}. Unlike pre-programmed trajectories, adaptive monitoring requires robots to determine where and when to collect data based on an evolving belief state \cite{154}. During early mission phases, uncertainty reduction may dominate, driving exploration of poorly characterized regions \cite{72}. As the mission progresses, however, priorities may shift toward maintaining coverage consistency, revisiting high-value locations, or stabilizing drift accumulation \cite{71}. Accordingly, belief-aware planners integrate uncertainty regulation directly into sampling strategies, treating localization stability, mapping confidence, or prediction variance as core planning variables rather than diagnostic afterthoughts \cite{64,65}.

At the same time, uncertainty regulation cannot be decoupled from resource sustainability. Long-duration deployments are inherently energy-limited, and propulsion, sensing, and computation compete for finite onboard resources \cite{97}. Therefore, adaptive sampling must be coordinated with energy-aware routing and speed scheduling \cite{155}. Current-assisted navigation, altitude modulation, and opportunistic loitering behaviors are often exploited to extend endurance while preserving data fidelity \cite{97,98}. In effect, monitoring transforms planning into a closed-loop stabilization problem in joint mission–belief–energy space.

Recent systems demonstrate how this integration is realized in practice. ElTobgui \textit{et al}.~\cite{46} propose SeaQuery, which fuses segmentation-derived ecological metrics with vision--language reasoning to support quantitative, context-aware reef assessment. Lin \textit{et al}.~\cite{89} introduce DREAM, a VLM-guided autonomy framework that improves coverage efficiency when target locations are unknown. Complementing decision-level advances, Alawode \textit{et al}.~\cite{130} develop AquaticCLIP to strengthen cross-modal grounding under perceptual degradation. Although differing in implementation, these systems share a common structural theme: perception quality, belief estimation, and motion decisions are co-designed under endurance constraints.

Against this background, persistent monitoring highlights that embodied intelligence must sustain epistemic coherence and resource stability over long horizons. Mission value, localization drift, and energy availability are dynamically intertwined. Therefore, this domain serves as a primary stress test for uncertainty-aware planning, long-duration control stability, and deployment-time reliability.

\subsection{Offshore Infrastructure Inspection and Interaction-Aware Autonomy}

In contrast to wide-area environmental monitoring, offshore infrastructure inspection operates under proximity-driven constraints, where safety, precision, and viewpoint geometry become dominant design factors \cite{5}. Whereas monitoring primarily stresses long-horizon belief stabilization and endurance, inspection places immediate emphasis on short-horizon loop reliability under tight spatial margins \cite{131}. Consequently, this domain exposes a distinct yet equally critical structural tension within embodied autonomy: perception quality and dynamic safety are inseparably coupled (Fig.~\ref{fig:4}b).

To understand this coupling, it is first necessary to consider the dynamic environment near subsea structures \cite{143}. Close-range maneuvering amplifies hydrodynamic disturbances due to flow separation, vortex shedding, and structural interference \cite{156}. These effects may induce unpredictable forces that destabilize the vehicle. As a result, inspection planning cannot be decoupled from dynamic feasibility \cite{82}. Safe stand-off distances must be maintained even while the vehicle approaches surfaces to collect high-resolution data \cite{143}. Moreover, because the operating envelope is narrow, minor control errors can rapidly propagate into collision risks. Stability margins and real-time disturbance rejection, therefore, become central to mission success \cite{156}.

At the same time, inspection tasks are inherently perception-driven. Unlike broad-area coverage missions, inspection requires specific viewpoints, incidence angles, and stand-off distances to ensure adequate feature visibility \cite{142,143}. Therefore, trajectory generation must explicitly incorporate sensing geometry as an optimization variable. The robot must position itself to maximize defect detectability while remaining within dynamically stable and controllable regimes \cite{157}. This creates a direct trade-off: improving sensing resolution may demand aggressive maneuvering, whereas conservative stabilization strategies may degrade inspection fidelity.

Beyond geometric considerations, semantic prioritization introduces an additional layer of coupling. Many inspection missions involve identifying corrosion, cracks, biofouling, or other defect signatures that require contextual interpretation \cite{75,158}. Language-based and vision–language models increasingly assist in mapping sensory input to semantic hypotheses and generating structured subgoals \cite{131,132}. However, semantic intent alone does not ensure physical feasibility. Without embodiment-aware constraints, high-level goals may produce trajectories that violate safety envelopes or exceed actuation limits.

For this reason, hierarchical autonomy architectures are commonly adopted. In such systems, semantic reasoning operates at an abstract supervisory layer, proposing inspection targets or priority regions, while lower-level planners and controllers enforce hydrodynamic feasibility, collision avoidance, and uncertainty regulation \cite{88,89}. This layered design ensures that semantic abstraction remains grounded in physical dynamics. For example, Akram \textit{et al}.~\cite{131} propose AquaChat, which translates natural language inspection commands into executable motion plans with event-triggered replanning. Similarly, Khorrambakht \textit{et al}.~\cite{133} develop a structured multi-agent inspection framework that integrates LLM-based supervision with DRL-based control, maintaining feasibility under dynamic disturbances.

Taken together, offshore inspection highlights a complementary embodiment principle: semantic reasoning, perception geometry, and dynamic safety must co-evolve continuously within a tightly regulated closed loop. Unlike monitoring, where the dominant challenge lies in long-term uncertainty accumulation, inspection stresses real-time stability and proximity-aware reliability. Consequently, this domain underscores the necessity of hierarchical yet tightly grounded embodied architectures in safety-critical underwater environments.

\subsection{Long-Horizon Exploration, Mapping, and Target-Seeking}

Building upon the monitoring and inspection scenarios discussed above, long-horizon exploration and mapping further intensify the structural couplings inherent in underwater embodied intelligence \cite{42, 159}. Whereas monitoring primarily regulates uncertainty under endurance constraints and inspection stresses proximity-driven safety, exploration combines unknown terrain, intermittent localization, and prolonged communication gaps within a single mission profile \cite{141}. In this setting, uncertainty does not merely fluctuate; instead, it accumulates over time. Consequently, exploration exposes one of the most systemic manifestations of constraint-coupled autonomy: state evolution, belief drift, and energy depletion interact over extended temporal scales (Fig.~\ref{fig:4}c).

To appreciate this challenge more concretely, consider the absence of reliable global positioning in many exploration missions \cite{162}. Dead reckoning errors accumulate continuously, acoustic localization updates may be sparse or delayed, and environmental disturbances often remain poorly characterized \cite{160, 161}. Under such conditions, purely geometric frontier expansion becomes insufficient. Instead, exploration must be belief-aware. Planners must dynamically balance outward expansion with revisitation strategies that constrain localization drift and maintain map consistency \cite{64,65}. In effect, mapping becomes a process of regulating epistemic growth rather than simply maximizing spatial coverage.

At the same time, terrain uncertainty introduces additional layers of dynamic complexity \cite{163}. Multimodal sensing, which often combines sonar, inertial measurements, and occasionally optical inputs, must be effectively fused to maintain reliable traversability estimation under conditions of degraded visibility \cite{144}. Because hydrodynamic conditions may vary with depth, seabed topology, and structural proximity, motion policies must remain adaptable to evolving flow regimes \cite{164}. As a result, exploration planning becomes a continuous negotiation among information gain, controllability, and safety margins \cite{165}. Aggressive frontier pursuit may accelerate coverage, yet it may also push the vehicle into dynamically unstable regions or exacerbate localization error accumulation \cite{166}.

Furthermore, in target-seeking missions, exploration extends beyond geometric mapping into semantic abstraction. Robots may be tasked with identifying ecological features, submerged objects, or non-cooperative moving targets \cite{167}. In such cases, language-based and vision–language models can provide contextual reasoning and memory, enabling adaptive refinement of search objectives under uncertain observations \cite{50,81}. However, as in inspection scenarios, semantic reasoning must remain grounded in dynamically feasible motion and energy limits. Without embodiment-aware filtering, high-level goals may conflict with stability envelopes or endurance constraints \cite{89}.

Representative systems demonstrate how these embodied principles are operationalized in practice. Girdhar \textit{et al}.~\cite{1} present CUREE, a field-deployed platform that integrates adaptive behaviors with multimodal perception for coral-reef investigation, illustrating sustained decision-making under real ocean uncertainty. Meng \textit{et al}.~\cite{81} develop UTracker, which distills simulated expert behavior into a visuomotor policy capable of robust tracking under dynamic perturbations. Ma \textit{et al}.~\cite{50} propose LEGO, an LLM-assisted optimization framework for underwater image restoration without ground-truth supervision, thereby enhancing perceptual reliability in low-visibility environments. Although these systems differ in emphasis, they share a common structural pattern: perception, belief regulation, and motion feasibility are jointly optimized rather than sequentially layered.

Collectively, long-horizon exploration reveals that autonomy is not merely path generation but the sustained regulation of epistemic stability under evolving environmental uncertainty. As mission duration increases, cross-layer error propagation becomes more pronounced: localization drift biases planning decisions; biased plans induce dynamic strain; dynamic instability accelerates energy depletion and sensing degradation. Therefore, exploration missions underscore the necessity of embodied autonomy operating explicitly within joint state–belief–energy space.

Moreover, as exploration scales from single-robot deployments to cooperative teams, communication sparsity and decentralized coordination introduce additional constraints. This progression naturally motivates the next domain, where scalability and cross-agent belief consistency become central design challenges.

\subsection{Communication-Constrained Multi-Robot Cooperation}

Building upon the long-horizon exploration scenarios discussed above, scaling from single-robot autonomy to cooperative multi-robot systems introduces an additional and fundamentally distinct layer of constraint coupling \cite{90}. While individual platforms must regulate state, belief, and energy internally, cooperative systems must simultaneously manage cross-agent consistency under severe communication limitations \cite{125}. In underwater environments, acoustic links provide low bandwidth, high latency, and intermittent connectivity \cite{91,92}. Consequently, coordination cannot rely on centralized, high-frequency information exchange. Instead, scalability becomes inseparable from communication physics itself \cite{96} (Fig.~\ref{fig:4}d).

To appreciate the implications of this limitation, consider how communication sparsity fragments shared belief states \cite{168}. Localization updates, map information, and task assignments may propagate asynchronously across agents, leading to inconsistent world models \cite{169}. As a result, centralized optimal planning becomes both impractical and fragile \cite{40}. Rather than maximizing instantaneous global optimality, embodied multi-robot systems increasingly prioritize decentralized robustness, ensuring that each agent can act coherently based on local observations even when global synchronization is incomplete \cite{115}.

In response to this structural constraint, modern cooperative frameworks adopt sparse and adaptive communication strategies \cite{171}. Agents must determine not only how to act locally, but also when and what to transmit. Communication is thus elevated from a passive data channel to an actively managed resource, trading coordination benefit against bandwidth and energy expenditure \cite{93}. Leader–follower structures, dynamic role switching, and relay behaviors are frequently employed to maintain functional connectivity while minimizing transmission load \cite{94,95}. Through such mechanisms, teams preserve collective performance despite fragmented information \cite{170}.

At the algorithmic level, multi-agent reinforcement learning (MARL) has emerged as a prominent approach for learning decentralized coordination policies \cite{172}. However, naive centralized-training assumptions often overlook real acoustic constraints \cite{190}. Consequently, recent methods explicitly integrate communication sparsity into policy design and optimization. Zhu \textit{et al}.~\cite{37} introduce LLM-EHT, synthesizing offline data to accelerate convergence under bandwidth limitations. Zhu \textit{et al}.~\cite{88} extend this framework with LLM-HMT to support multi-task AUV clusters via hybrid online–offline training. Wang \textit{et al}.~\cite{134} propose MACSAC, a continuous-space multi-agent SAC framework with probabilistic target modeling and adaptive role allocation, enabling efficient coordination without dense communication. These approaches collectively demonstrate a shift from communication-heavy coordination toward decentralized resilience.

Crucially, communication constraints are not purely informational; they also impose energetic costs. Acoustic transmission consumes power, and sustained communication reduces mission endurance \cite{96}. Therefore, coordination strategies must balance synchronization precision against energy sustainability \cite{174}. In long-duration deployments, conservative communication scheduling may preserve endurance but slow convergence, whereas frequent updates may improve coordination at the expense of operational lifespan \cite{173}. Embodied autonomy must therefore negotiate trade-offs among coordination accuracy, energy expenditure, and robustness to dropout or delay.

In summary, communication-constrained cooperation reveals an additional dimension of embodiment: collective intelligence is bounded by physical communication limits. Sparse bandwidth and delayed updates fragment shared belief states, making decentralized resilience more critical than centralized optimality. As system scale increases, robustness emerges not from increasing algorithmic complexity alone, but from bounding cross-agent error propagation under communication and energy constraints.

In this sense, multi-robot cooperation represents the culmination of constraint-coupled autonomy. It integrates uncertainty regulation, dynamic feasibility, semantic abstraction, and resource sustainability across multiple interacting agents. The application domains reviewed in this section collectively illustrate that underwater embodied intelligence is not defined by isolated algorithmic advances, but by its ability to internalize physical constraints across perception, planning, and control layers, while maintaining coordination under communication and energy constraints.

\section{Challenges and Research Outlook}
Although the preceding application domains demonstrate the potential of embodied autonomy, they also expose recurring structural limitations. We therefore now turn to the fundamental challenges that continue to constrain underwater embodied intelligence. While the preceding sections demonstrated how embodied principles enable progress under real ocean conditions, they also implicitly revealed recurring sources of fragility. These limitations are not isolated technical shortcomings that can be resolved independently. Rather, they reflect interconnected failure modes that propagate across perception, planning, and control layers.

Indeed, as underwater systems become more adaptive and integrated, cross-layer coupling becomes stronger. Errors in perception influence planning objectives; planning decisions reshape dynamic feasibility; control instability amplifies uncertainty; and coordination breakdown further fragments shared belief states. When these couplings are insufficiently regulated, small deviations may cascade across layers and accumulate over long horizons.

Accordingly, instead of enumerating challenges in isolation, we organize current limitations through a cross-layer failure taxonomy. By clarifying how uncertainty, instability, and coordination constraints interact, we aim to outline principled research directions toward robust, scalable, and verifiable underwater embodied systems.

\subsection{Environmental Uncertainty and Distribution Shift}

We begin with environmental uncertainty, as it forms the foundational constraint underlying all subsequent failure modes \cite{175}. In underwater environments, non-stationarity is the rule rather than the exception. Ocean currents vary with depth, terrain, and seasonal conditions; hydrodynamic coefficients shift with operating regimes; and sensing quality degrades due to turbidity, acoustic multipath, and intermittent localization \cite{137,119,38}. Consequently, both model-based and learning-based autonomy systems frequently encounter conditions that deviate from their original design assumptions.

This variability becomes particularly critical in learning-enabled frameworks. Policies trained in simulation or under limited field trials may exhibit strong performance within their training distribution, yet behave unpredictably when exposed to unseen disturbances \cite{106}. Although domain randomization, belief-aware planning, and online adaptation can partially reduce this gap, rare but high-impact events, such as abrupt current transitions, sensor drift, actuator degradation, or communication dropouts, remain inherently difficult to anticipate and model in advance \cite{135,72,136}. Thus, distribution shift represents not a marginal issue, but a structural property of marine environments.

Moreover, increasing model complexity alone does not guarantee improved robustness \cite{118}. Larger neural networks or richer training datasets may capture more variability, but without an explicit representation of epistemic uncertainty, learned policies may still produce overconfident and unsafe decisions under novel conditions \cite{176, 177}. Over extended missions, even small modeling discrepancies may accumulate into significant trajectory drift, belief inconsistency, or energy inefficiency \cite{178,179}. In this way, environmental uncertainty does not merely affect instantaneous performance; it reshapes long-horizon autonomy dynamics.

Therefore, the core challenge extends beyond refining environmental models. Instead, uncertainty quantification must be embedded directly into autonomy loops. Future systems should propagate confidence measures through perception, planning, and control layers, enabling adaptive and conservative decision-making under incomplete knowledge. In other words, robustness must be treated as a primary design objective rather than an emergent byproduct of increased data scale or algorithmic sophistication.

\subsection{Closed-Loop Reliability, Safety, and Verifiability}

While environmental uncertainty defines the external variability faced by underwater systems, closed-loop reliability concerns how such uncertainty propagates within the autonomy stack itself. As discussed in the preceding subsection, distribution shift introduces epistemic bias at the perception and estimation levels \cite{180,181}. However, the consequences of this bias rarely remain localized. Instead, because underwater autonomy operates through tightly coupled feedback loops, small estimation errors may gradually amplify across planning and control layers.

To clarify this amplification mechanism, consider the recursive structure inherent in embodied autonomy \cite{47}. Perception outputs shape state estimation; estimated states define planning objectives; planning decisions generate control commands; and control actions, in turn, reshape future sensing conditions \cite{36,183}. In long-duration missions, even minor deviations in the early stages may accumulate into significant localization drift, unstable maneuvers, or inefficient energy usage \cite{182}. Therefore, reliability must be evaluated not only by instantaneous accuracy, but also by the stability of feedback regulation over extended horizons.

In response to this challenge, hybrid control architectures have emerged as a pragmatic compromise. By combining classical stabilizing controllers with learning-based residual policies, such systems attempt to improve disturbance rejection while preserving conservative behavior in nominal regimes \cite{111,118}. This layered design helps mitigate catastrophic failures under moderate disturbances. Nevertheless, empirical robustness does not automatically translate into formal guarantees. Under partial observability, time-varying hydrodynamics, and actuator saturation, rigorous stability proofs for learning-augmented systems remain limited \cite{69,134}.

Moreover, as autonomy stacks incorporate higher levels of abstraction, additional reliability concerns arise. The integration of language-based and vision–language models enables semantic reasoning and flexible task interpretation \cite{131,46}. Yet, this abstraction layer introduces uncertainty in goal formulation and subgoal generation. Without explicit embodiment-aware constraints, semantically plausible outputs may lead to dynamically infeasible trajectories or unsafe control commands \cite{100}. Thus, semantic reasoning must be grounded within physically realizable motion and verified against safety envelopes before execution.

For these reasons, verifiability has emerged as a central research frontier \cite{123}. Rather than viewing learning and control as separate paradigms, future architectures must integrate adaptation within provably safe operational bounds. Promising directions include Lyapunov-constrained reinforcement learning, passivity-preserving residual control, control barrier functions embedded within learned planners, and runtime monitoring mechanisms capable of supervising policy outputs in real time \cite{185,186,187}. These approaches aim to ensure that adaptation remains bounded within stability-preserving regions.

In practice, verifiability is not merely a theoretical concern. In safety-critical inspection, cooperative exploration, or infrastructure monitoring missions, small control deviations may escalate into collisions, entanglement, or mission failure \cite{188,189}. Therefore, structured fallback mechanisms, interpretable control layers, and uncertainty-aware supervision are essential components of trustworthy deployment.

Overall, closed-loop reliability highlights that robustness must be assessed dynamically rather than statically. The challenge extends beyond minimizing tracking error under nominal conditions; it requires regulating feedback amplification under uncertainty and abstraction. Addressing this challenge will demand deeper integration between control theory, uncertainty quantification, and learning-based adaptation within unified embodied architectures.

\subsection{Scalability, Coordination, and Resource-Constrained Deployment}

Building upon the preceding discussions of environmental uncertainty and closed-loop reliability, a third structural challenge becomes evident as underwater systems increase in scale and complexity: resource-constrained deployment. Whereas Sections IV-A and IV-B focused on uncertainty propagation within individual autonomy loops, scalability introduces cross-agent interactions and cross-resource trade-offs that further intensify system coupling.

To start with, scaling from single-robot prototypes to persistent multi-robot deployments fundamentally reshapes coordination requirements \cite{146}. Underwater acoustic communication provides limited bandwidth, high latency, and intermittent connectivity \cite{125,139}. As a result, belief updates, map synchronization, and task allocation decisions propagate asynchronously across agents \cite{86}. Consequently, maintaining centralized coordination becomes increasingly impractical as the system size grows \cite{40}. Instead of pursuing globally optimal coordination at every timestep, scalable autonomy must prioritize decentralized robustness under communication sparsity.

At the same time, communication constraints are tightly intertwined with energy limitations \cite{190}. Propulsion, sensing, communication, and onboard computation compete for finite power budgets \cite{191}. In long-duration deployments, excessive communication or computational overhead may directly shorten mission lifespan. Therefore, scalability cannot be achieved simply by increasing algorithmic sophistication or adding more coordination layers. Rather, intelligence must be co-designed with energy sustainability from the outset \cite{54}.

\begin{figure*}[!t]
    \centering
    \includegraphics[width=1.0\linewidth]{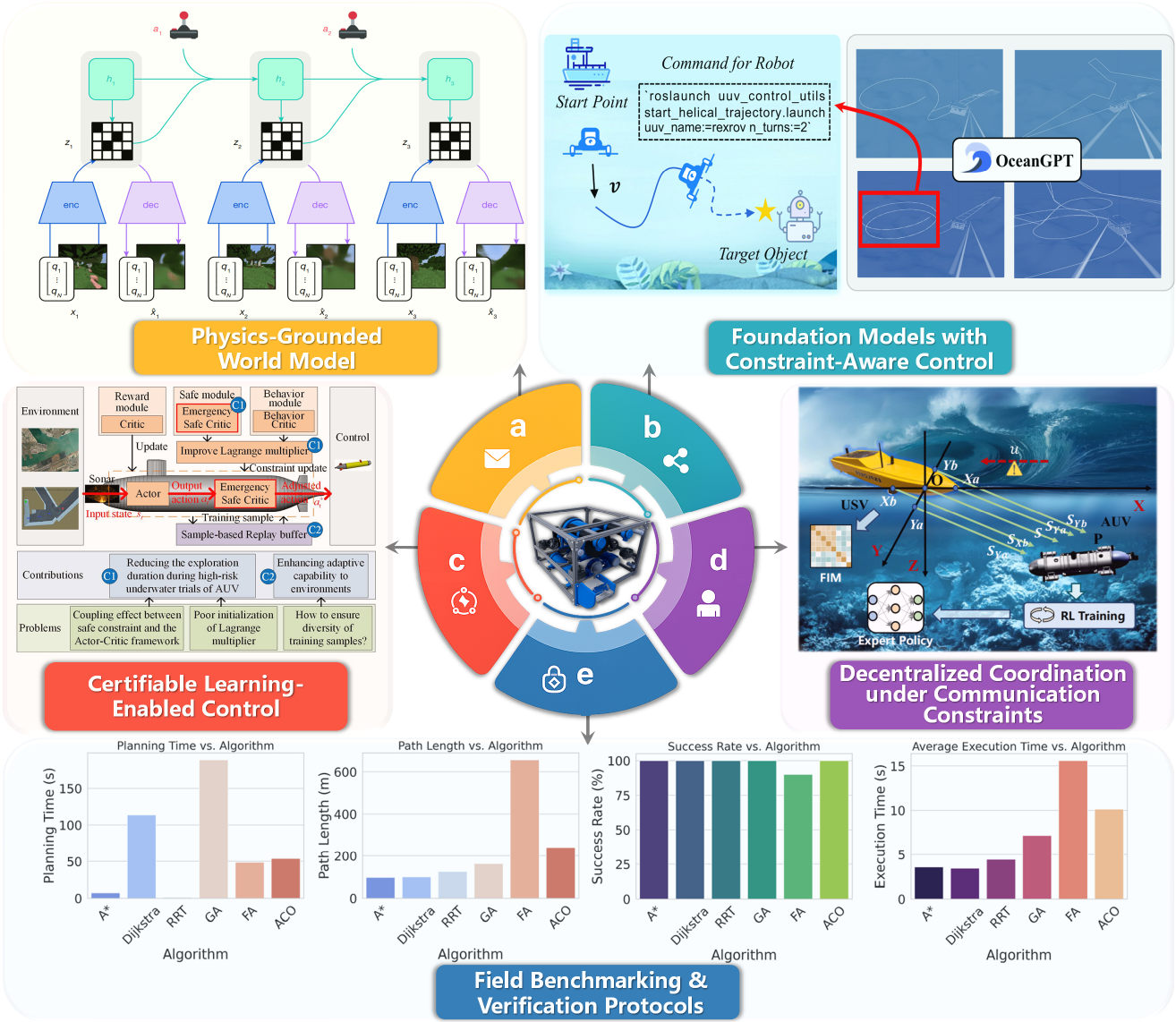}
    \caption{\textbf{Emerging research directions for resilient underwater embodied intelligence}. Future underwater autonomy requires integrated strategies that regulate cross-layer coupling across perception, planning, and control, while explicitly accounting for communication, coordination, and deployment constraints. (a) Dreamer introduces a general model-based reinforcement learning algorithm that learns latent world models of environment dynamics and improves behaviour by imagining future trajectories, illustrating how physics-grounded world models can support anticipatory planning under uncertainty \cite{198}. (b) OceanGPT presents a domain-specific large language model for ocean science that integrates heterogeneous oceanographic data and instruction-generation frameworks, highlighting the potential of foundation models grounded in ocean-domain knowledge \cite{42}. (c) The Leash Actor–Critic (LAC) method proposes a constraint reinforcement learning framework that combines an emergency safety critic with Lagrangian optimization to restrict unsafe actions during AUV path planning, demonstrating how stability and safety constraints can be embedded within learning-enabled control \cite{138}. (d) A collaborative USV–AUV system integrates Fisher-information–based localization with reinforcement-learning-based cooperative planning, illustrating scalable decentralized coordination under communication and environmental constraints \cite{118}. (e) An open-source AUV path-planning benchmarking platform built on realistic underwater simulation environments provides standardized evaluation scenarios and metrics, highlighting the importance of field benchmarking and reproducible validation for underwater autonomy \cite{199}.}
    \label{fig:6}
\end{figure*}

Furthermore, the growing integration of large foundation models introduces additional computational burdens \cite{42}. While such models enhance semantic abstraction and task flexibility, they also increase inference latency and power consumption \cite{130}. In underwater contexts, where cloud connectivity is unreliable and onboard processors are constrained, these computational demands become critical factors for deployment \cite{45}. Thus, scaling intelligence requires careful balancing between representational richness and real-time feasibility.

Beyond hardware and computation, scalability also complicates validation and reproducibility \cite{135}. As system complexity increases, emergent behaviors and rare failure modes become harder to isolate and replicate \cite{15}. Unlike simulation benchmarks, real ocean deployments expose compounding disturbances that may not appear in controlled trials \cite{88}. The absence of standardized field evaluation protocols further obscures systematic comparison across platforms and missions.

Taken together, scalability, coordination, and resource constraints reveal that embodied autonomy must remain deployment-aware at every design stage. Robust multi-robot systems will not emerge solely from larger models or more complex optimization schemes. Instead, they require decentralized coordination strategies, adaptive communication scheduling, energy-aware planning, and computationally efficient architectures that internalize physical constraints.

In this sense, scalability does not represent a separate category of difficulty, but rather an amplification of the uncertainties and feedback couplings discussed in earlier subsections. As system scale increases, environmental variability, feedback instability, and communication sparsity interact more strongly, creating cascading risks across layers. This observation naturally motivates a unified cross-layer failure taxonomy, which we introduce next.

\subsection{A Cross-Layer Failure Taxonomy}

Having examined environmental variability, feedback instability, and scalability constraints in isolation, we now synthesize these challenges into a unified structural perspective. Although each category appears distinct when viewed independently, they share a deeper commonality: in embodied underwater systems, failures rarely remain confined to a single layer. Instead, errors propagate across perception, planning, and control layers, often amplifying as they cascade over time.

To make this interaction explicit, we organize failure modes into three interconnected layers:

(i) \textbf{Epistemic failures}, arising from partial observability, belief fragmentation, and distribution shift;

(ii) \textbf{Dynamic failures}, induced by hydrodynamic instability, actuator saturation, unsafe control responses, or poorly grounded planning outputs;

(iii) \textbf{Coordination failures}, caused by communication sparsity, asynchronous updates, and inconsistent shared world models.

At first glance, this taxonomy may resemble a modular decomposition. However, such a view would be misleading. In underwater embodied systems, these layers are tightly coupled through recursive feedback. For instance, epistemic failures distort state estimates and confidence measures, thereby biasing planning objectives \cite{83}. Distorted plans may push controllers toward aggressive or unstable operating regimes, increasing the likelihood of dynamic failure \cite{79}. In turn, unstable dynamics can degrade sensing quality, accelerate energy depletion, or trigger emergency maneuvers, thereby amplifying epistemic uncertainty \cite{144}. When multiple agents are involved, these disturbances may propagate across communication channels, leading to coordination breakdown and belief inconsistency \cite{72}.

Crucially, this cascade structure implies that addressing failures at a single layer may be insufficient. Improving perception accuracy without regulating control stability may still result in unsafe behavior under unexpected disturbances \cite{27}. Conversely, enforcing conservative control without resolving belief fragmentation may reduce efficiency or compromise mission objectives \cite{69}. Therefore, robust autonomy cannot be achieved by optimizing perception, planning, and control subsystems independently. Instead, it requires bounding cross-layer error propagation across the entire embodied stack.

Furthermore, the severity of cascading failures increases with mission duration and system scale. In long-horizon deployments, small epistemic biases may accumulate into substantial trajectory drift or map inconsistency \cite{57}. In multi-robot settings, asynchronous belief updates may magnify local dynamic instabilities into system-wide coordination disruptions \cite{168}. Thus, failure severity is determined not only by instantaneous error magnitude, but also by the system’s capacity to regulate amplification over time.

Collectively, this cross-layer failure taxonomy reframes underwater embodied intelligence as a stability problem under coupled uncertainty. The central design question therefore shifts from “How accurate is each module?” to “How does uncertainty propagate across modules, and how can amplification be bounded?” Addressing this question will require integrated design strategies that simultaneously regulate epistemic confidence, dynamic feasibility, and coordination consistency within unified embodied architectures.

\subsection{Future Directions}

Having clarified the cascading structure of failures in underwater embodied systems, we now shift from diagnosis to prescription. If autonomy breakdown arises from cross-layer amplification of epistemic uncertainty, dynamic instability, and coordination mismatch, then future progress cannot rely on isolated algorithmic refinements alone. Instead, it must be grounded in integrated, constraint-aware design strategies that explicitly regulate cross-layer coupling.

At the outset, advancing physics-grounded world models represents a foundational direction \cite{198} (Fig.~\ref{fig:6}a). In many current learning-based systems, environmental variability is treated primarily as a stochastic disturbance \cite{119}. However, ocean dynamics often exhibit structured spatial and temporal regularities that can, in principle, be modeled probabilistically \cite{193}. Therefore, future autonomy frameworks should integrate hydrodynamic modeling, sensing degradation processes, and energy consumption patterns within unified predictive representations. By coupling physical priors with belief-aware inference, such world models may enable planning that is not merely reactive to disturbances, but anticipatory under uncertainty \cite{192}.

At the same time, the integration of scalable foundation models with constraint-aware control architectures requires principled grounding mechanisms \cite{42} (Fig.~\ref{fig:6}b). As discussed earlier, semantic abstraction enhances flexibility and task generalization; yet, it also risks decoupling high-level reasoning from dynamic feasibility \cite{194}. To reconcile this tension, intermediate structured representations, such as feasibility-filtered subgoal graphs, constrained task hierarchies, or uncertainty-aware goal proposals, may serve as effective bridges between abstract reasoning and low-level stabilization \cite{183}. Through such mechanisms, abstraction can enhance adaptability without destabilizing closed-loop execution.

Beyond representation and abstraction, certifiable learning-enabled control remains a critical frontier \cite{138} (Fig.~\ref{fig:6}c). Although hybrid architectures have improved empirical robustness, formal guarantees under partial observability and distribution shift remain limited \cite{196}. Consequently, embedding stability analysis directly into learning loops becomes essential. Approaches such as Lyapunov-constrained reinforcement learning, passivity-preserving residual control, control barrier functions integrated with learned planners, and energy-aware adaptive regulation offer promising pathways \cite{185,187,195}. By enforcing stability-preserving constraints during policy optimization, learning can shift from opportunistic performance maximization toward bounded, reliable adaptation.

Furthermore, scalability demands renewed attention to decentralized coordination under communication constraints \cite{118,168} (Fig.~\ref{fig:6}d). Rather than assuming reliable synchronization, future multi-robot systems should incorporate asynchronous belief updates, communication sparsity, and limited bandwidth directly into their optimization objectives \cite{139}. Learning when to communicate, compressing information intelligently, and maintaining local consistency under delayed updates are not peripheral considerations, but core design elements for large-scale underwater deployments \cite{118}.

In addition to algorithmic and architectural advances, systematic field benchmarking and verification protocols are urgently needed \cite{130}. While simulation environments enable rapid iteration, real ocean deployments expose compounding disturbances and rare failure modes that challenge reproducibility \cite{111,199} (Fig.~\ref{fig:6}e). Without shared datasets, stress-test scenarios, and cross-platform evaluation standards, progress may remain fragmented. Establishing community-wide benchmarks would, therefore, strengthen comparability and accelerate collective advancement \cite{176}.

In summary, these research directions converge on a unifying insight: the next phase of underwater embodied intelligence will not be defined solely by model scale or computational capacity. Rather, progress will hinge on tight co-design across ocean engineering, control theory, and artificial intelligence. By internalizing physical embodiment, uncertainty representation, communication constraints, and resource sustainability within unified autonomy architectures, future systems may achieve robustness that extends beyond laboratory validation into sustained real-world operation.

\subsection{Emerging Design Principles}

Building upon the research directions outlined above, we now distill a set of design principles that emerge consistently across the application domains, the failure taxonomy, and the future directions discussed above. Rather than prescribing specific algorithms, these principles articulate structural guidelines for constructing robust underwater embodied systems under tightly coupled physical and informational constraints.

To begin with, learning should operate within physical envelopes rather than replace physics-based control. As repeatedly emphasized throughout this review, hydrodynamic coupling, actuator limits, buoyancy-induced restoring forces, and energy constraints fundamentally shape feasible behavior. Therefore, learning modules should augment, adapt, and refine physics-informed controllers while preserving stability-preserving structure. By embedding adaptation within known dynamic limits, autonomy can remain both flexible and bounded.

Closely related to this principle, sensing must be treated as an actively controllable resource rather than a passive input stream. Because perception quality depends on motion, viewpoint geometry, environmental conditions, and communication availability, planning and control decisions directly influence epistemic stability. Consequently, embodied systems should explicitly regulate information acquisition by coupling motion strategies with uncertainty reduction objectives. In doing so, belief evolution becomes a controllable variable rather than an uncontrolled byproduct of movement.

Furthermore, uncertainty and communication sparsity must be explicitly represented within autonomy loops. Instead of relying solely on point estimates or assuming synchronous coordination, systems should propagate confidence measures across perception, planning, and coordination layers. This is particularly critical in multi-robot deployments, where fragmented belief states may otherwise destabilize cooperation. Designing policies that remain coherent under delayed or partial information exchange is therefore essential for scalable autonomy.

At a higher level of abstraction, semantic reasoning and low-level stabilization should remain hierarchically separated yet tightly grounded. While foundation models and high-level planners enable flexible task interpretation and semantic generalization, their outputs must be filtered through constraint-aware planning modules and safety-enforcing control layers. Maintaining this structured separation preserves interpretability and prevents abstract reasoning from bypassing dynamic feasibility constraints.

In addition, intelligence and energy must be co-optimized under mission constraints. Computational overhead, communication load, aggressive maneuvering, and sensing operations all consume limited onboard resources. As a result, efficiency is not merely a secondary performance metric, but a structural requirement for sustained deployment. Autonomy architectures must therefore balance decision sophistication against endurance preservation.

Overall, these principles suggest a broader reframing of underwater embodied intelligence. Rather than viewing progress as the accumulation of increasingly complex models, it may be more productive to view autonomy as the disciplined regulation of interaction within a dynamically coupled marine medium. In this perspective, intelligence emerges from continuous negotiation among uncertainty, controllability, communication limitations, and resource sustainability.

Ultimately, the success of future underwater robotic systems will depend not only on algorithmic innovation, but on how tightly perception, planning, control, communication, and energy management are integrated within physically grounded architectures. By internalizing constraint coupling rather than treating it as an external disturbance, underwater embodied systems may achieve autonomy that is not only adaptive, but resilient, scalable, and verifiable under real ocean conditions.


\bibliographystyle{IEEEtran}
\bibliography{IEEEexample}

\addtolength{\textheight}{-12cm}

\end{document}